\documentclass{article} 
\usepackage{iclr2023,times}

\usepackage{amsmath,amsfonts,bm}

\def\eqref#1{equation~\ref{#1}}

\def\1{\bm{1}}

\DeclareMathAlphabet{\mathsfit}{\encodingdefault}{\sfdefault}{m}{sl}
\SetMathAlphabet{\mathsfit}{bold}{\encodingdefault}{\sfdefault}{bx}{n}

\usepackage{hyperref}
\usepackage{url}
\usepackage{wrapfig}
\usepackage{graphicx}

\title{Exploring the Limits of Semantic Image\\ Compression at Micro-bits per Pixel}

\author{Jordan Dotzel\textsuperscript{1}\thanks{Equal contribution},\hspace{0.25em}
        Bahaa Kotb\textsuperscript{1}\footnotemark[1],\hspace{0.25em}
        James Dotzel\textsuperscript{2},\hspace{0.25em}
        Mohamed Abdelfattah\textsuperscript{1},\hspace{0.25em}
        Zhiru Zhang\textsuperscript{1} \\ 
        \textsuperscript{1}Cornell University, \textsuperscript{2}Penn State University 
}

\iclrfinalcopy

\begin{document}

\maketitle

\vspace{-20pt}
\begin{abstract}
Traditional methods, such as JPEG, perform image compression by operating on structural information, such as pixel values or frequency content. These methods are effective to bitrates around one bit per pixel (bpp) and higher at standard image sizes. In contrast, text-based semantic compression directly stores concepts and their relationships using natural language, which has evolved with humans to efficiently represent these salient concepts. These methods can operate at extremely low bitrates by disregarding structural information like location, size, and orientation. In this work, we use GPT-4V and DALL-E3 from OpenAI to explore the quality-compression frontier for image compression and identify the limitations of current technology. We push semantic compression as low as 100 $\mu$bpp (up to $10,000\times$ smaller than JPEG) by introducing an iterative reflection process to improve the decoded image. We further hypothesize this 100 $\mu$bpp level represents a soft limit on semantic compression at standard image resolutions. 
\end{abstract}

\vspace{-12pt}
\section{Introduction}
\label{sec:intro}
\vspace{-10pt}

\begin{wrapfigure}{r}{0.50\textwidth}
  \vspace{-30pt}
  \includegraphics[width=.50\textwidth]{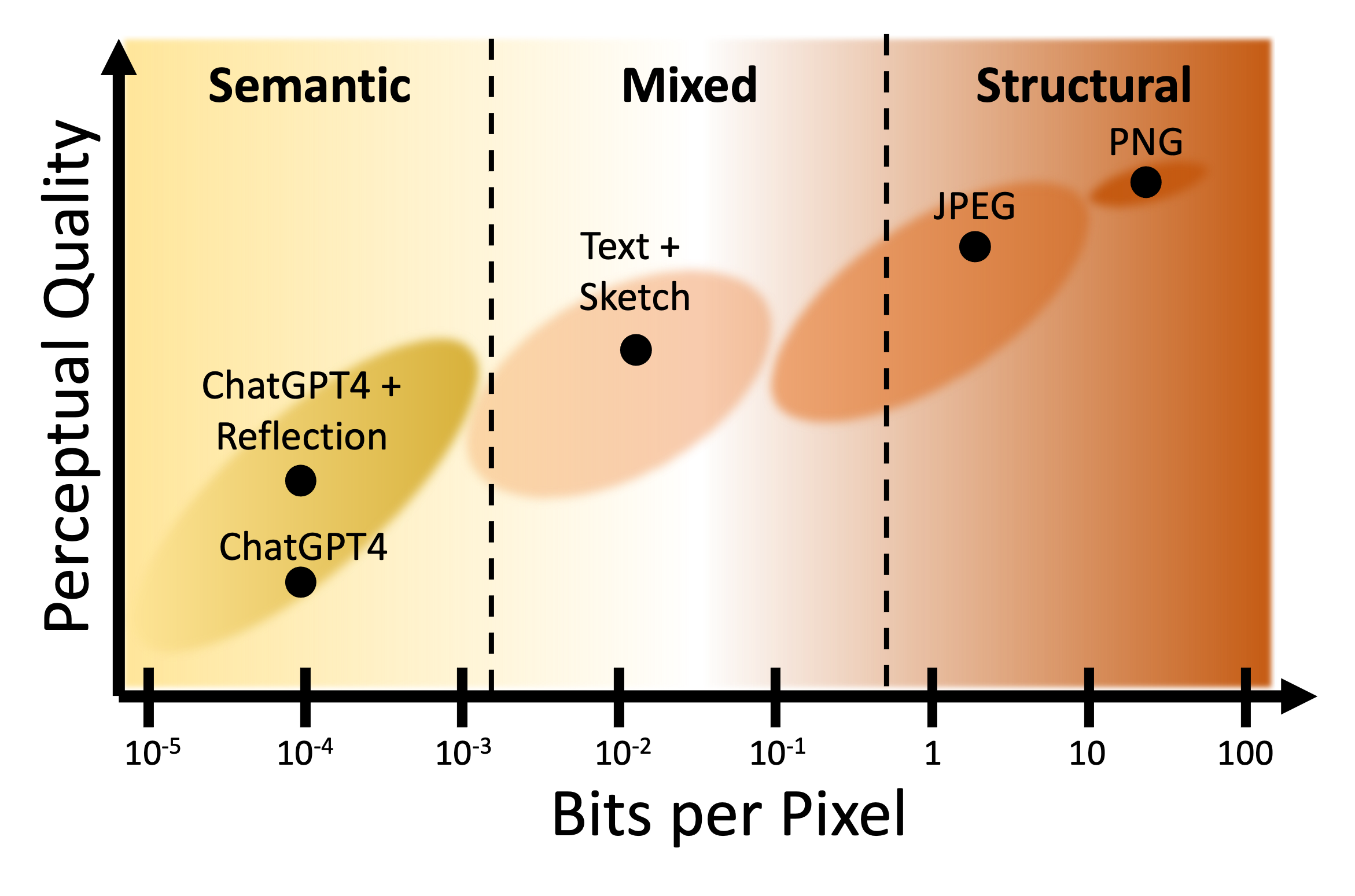}
  \vspace{-25pt}
  \caption{\textbf{Compression Regions}: This work explores the limits of semantic compression with ChatGPT4 and demonstrates improvements through iterative reflection.}
  \vspace{-5pt}
  \label{fig:regions}
\end{wrapfigure}

Modern image compression can be classified as lossless, where all information is preserved, or lossy, where some information is lost.
Lossy compression transfers data at lower bitrates at the cost of quality and typically attempts to adapt to human preferences by removing imperceptible or barely perceptual information . 
For example, JPEG is a lossy compression technique that uses a frequency transform to retain only the most human-perceptible frequencies.
Inspired by prior work~\cite{weissman2023textual}, we classify image compression into three major regions, as shown in Fig.~\ref{fig:regions}.
The structural region contains techniques that preserve the original pixel-level structure.
In this region, JPEG typically operates between $10^{-1}$ and 10 bits per pixel (bpp).
The semantic region, on the other hand, preserves only the human-centric, \textit{semantic} information, often dropping precise locations, sizes, orientations, and viewing angles.
We demonstrate that this region can achieve compression to hundreds of micro-bits per pixel ($\mu$bpp) by using natural language, which has co-evolved alongside civilization to efficiently capture the most important human-centric concepts.
Finally, the mixed region combines partial semantic and structural information.
For example, Text + Sketch uses text descriptions with high-density hierarchical sketches at $10^{-3}$ to $10^{-2}$ bpp~\cite{lei2023text}.

This work uses ChatGPT4 from OpenAI, built with GPT4-V (vision) ~\cite{openai2023gpt4} and DALL-E3~\cite{betker2023improving}, to explore the lower limits of the semantic region.
To adjust for limitations with these models, it additionally introduces \textit{image reflection}, adapted from the code generation literature~\cite{shinn2023reflexion}, to iteratively improve generated images.
This method is a proof-of-concept using public models and achieves bitrates up to $10,000\times$ smaller than JPEG by trading off precise structural information for low-resolution semantics.
It explores the limits of image compression with current technology, and suggests a practical limit around 100 $\mu$bpp at $1024\times1024$ resolution.
At even higher resolution, it scales better than traditional methods since semantic information grows sub-linearly with respect to image resolution.
This suggests that pure semantic compression with improved technology has future applications that require transmitting large amounts of high-resolution data across images, video, and 3D objects, especially within collaborative virtual worlds.

\section{Method}
\label{sec:method}
\vspace{-10pt}

\begin{figure}[t]
\vspace{-30pt}
  \centering
  \includegraphics[width=\textwidth]{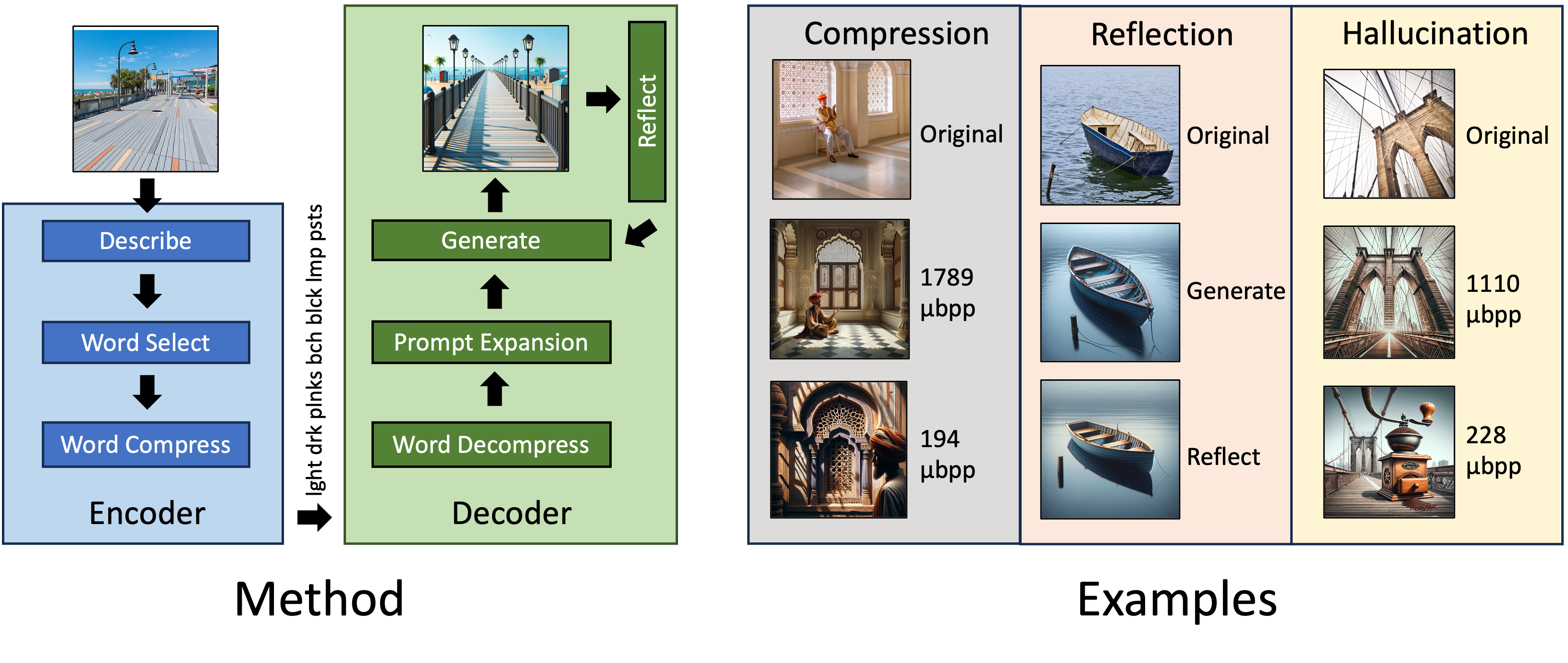}
  \vspace{-20pt}
  \caption{\textbf{Method and Examples:} ChatGPT4 can perform semantic compression at the 100 $\mu$bpp level, capturing only the most important concepts in the image with respect to human preferences. }
  \label{fig:method}
  \vspace{-15pt}
\end{figure}

We use GPT-4V as the base encoder and DALL-E3 as the base decoder since these models have demonstrated state-of-the-art language compression, image comprehension, and generation.
As shown in Fig~\ref{fig:method}, the encoder first analyzes the input image and extracts a very detailed description.
Then, it controls the compression level by selecting the $N$ most important words, ignoring punctuation, articles, and other unimportant elements.
It later further compresses these words by removing as many characters as possible and using only the most common fifteen consonants (along with the space character) to restrict each character to a four-bit representation.
Large language models like GPT-4V excel at this textual compression and decompression task~\cite{gilbert2023semantic} by holistically looking across the entire compressed description to decode each word.
Without explicit prompting, GPT-4V even swaps synonyms for words to reduce the character count and to avoid difficult and ambiguous word compression.

This compressed description is then passed to the decoder for reconstruction and expansion into a fuller, more natural prompt and passed to DALL-E3 to produce an image.
If there is sufficient context, typically over 500 $\mu$bpp, then a novel image reflection technique improves the image through iterative generation, description, and reflection.
This first produces a new detailed description of the generated image and then finds the most salient difference between this description and the original description.
Then, the DALL-E3 regenerates the image with the specified edit.
Higher bitrates have more context and can benefit from more iterations of reflection, yet we find typically only one or two reflections capture the most important issues.
All prompts are listed in Appendix~\ref{sec:prompts}.

\vspace{-12pt}
\section{Analysis}
\label{sec:analysis}
\vspace{-10pt}

Figure~\ref{fig:method} highlights examples of varying bitrates, reflection, and hallucinated features (more examples in Appendix~\ref{sec:examples}).
The first example shows the effects of progressive compression from 1789 $\mu$bpp to 194 $\mu$bpp, where the more compressed image loses information related to the sitting man, floor tiles, and his location.
It also demonstrates the impressive language capabilities of GPT-4V by correctly decompressing `ndn trdtnl rch wndw lt shdw' into `indian traditional arch window lattice shadow'.
At this image size, 100 $\mu$bpp (25 chars) is the practical limit, and high-quality reconstructions are only possible with the most common angles on the most common objects.
Within the same figure, the boat image demonstrates reflection at 800 $\mu$bpp by recognizing the differences in the boat stern shape and interior color and fixes both of these issues sequentially.
This example also shows that these models struggle with object orientation, and DALL-E3 currently cannot incorporate all the details or independently edit one part of the image without modifying another part.
Finally, the bridge image shows an example hallucination by decompressing `grndr' as grinder instead of grandeur at 118 $\mu$bpp.
It is also possible to reflect at this word decompression stage, since the model recognized the strangeness of this word when prompted, yet this would limit the diversity of images.

Despite these minor shortcomings, ChatGPT4 is a proof-of-concept for useful semantic compression that achieves near human-level ability in describing images, selecting the most important words, and compressing these representations.
Currently, the descriptive abilities of GPT-4V outperform the generative abilities of DALL-E3, and therefore improvements will come from increased sensitivity to more descriptive detail and the ability to perform precise, independent edits. 


\bibliography{iclr2023}
\bibliographystyle{iclr2023}

\newpage
\appendix

\section{Examples}
\label{sec:examples}

\begin{figure}[ht]
  \centering
  \includegraphics[width=\textwidth]{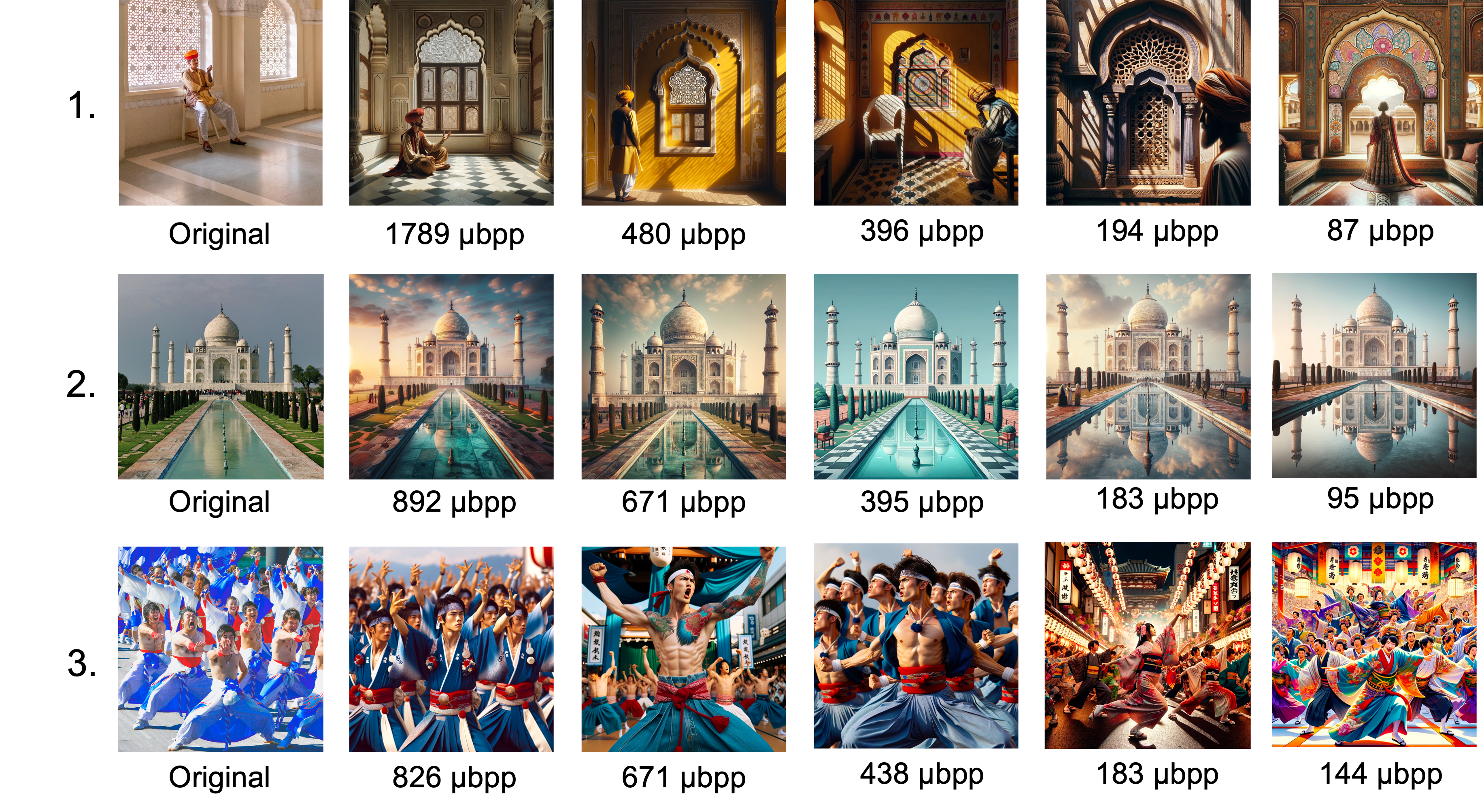}
  \includegraphics[width=\textwidth]{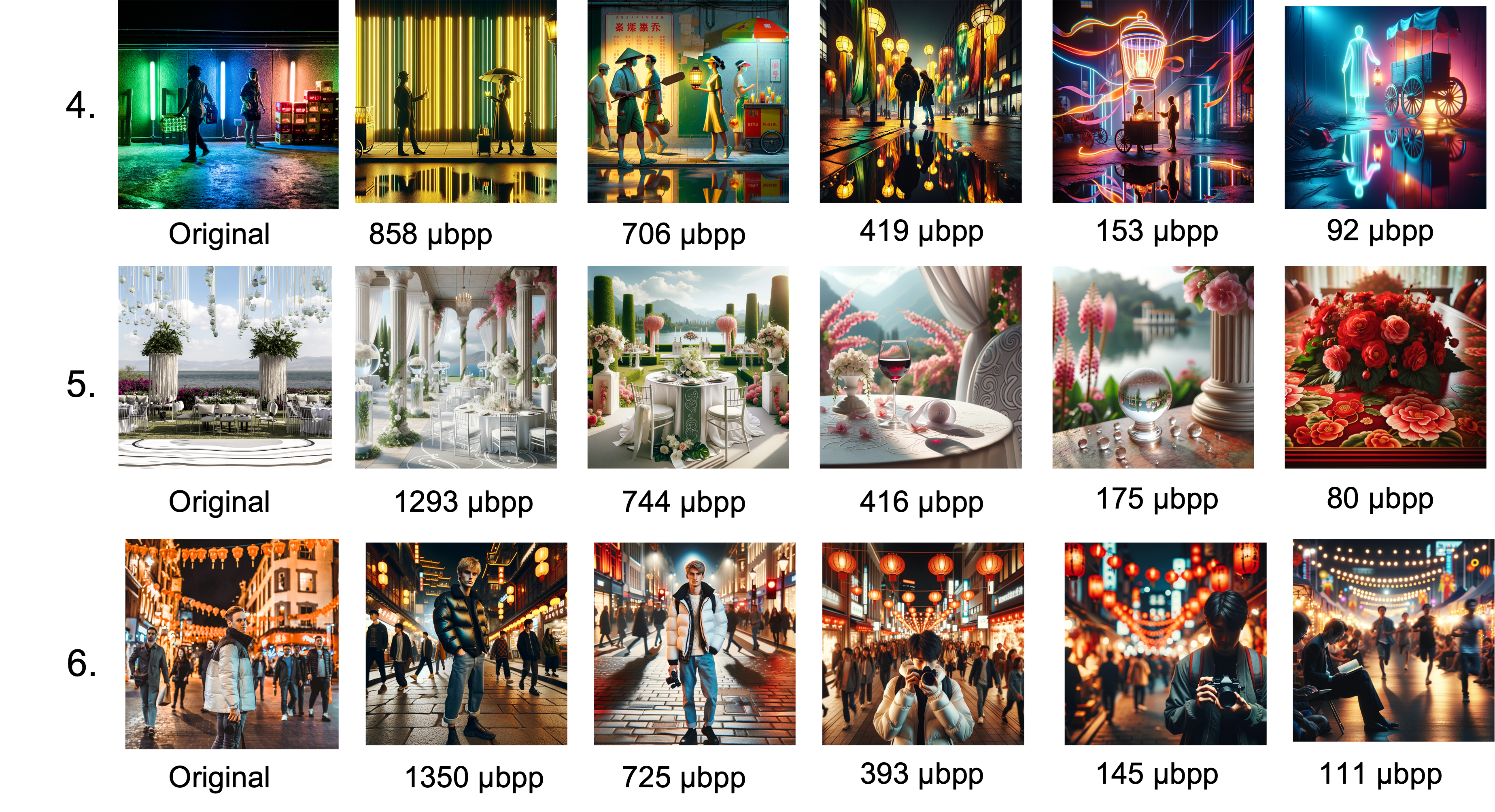}
  \vspace{-10pt}
  \caption{\textbf{Compression Examples:} The first example shows the progressive loss of contextual information from tile details, room color, location of the man, sitting vs. standing. The second example, on the other hand, shows that landmarks and proper nouns like the Taj Mahal taken from standard angles can be compressed extremely small to 10s of $\mu$bpp since a significant amount of information is captured within a few words. The third example again shows the gradual loss of context, color, gender, and location. The fourth example shows the progressive loss of contextual information including light colors, figure position, and style of the lights. The fifth example shows that with heavy compression it hyper-focuses on certain arbitrary details like the flowers. Finally, the last example shows the loss of information about the jacket color and other details with higher compression.}
  \label{fig:examples1}
\end{figure}

\begin{figure}[ht]
  \centering
  \includegraphics[width=\textwidth]{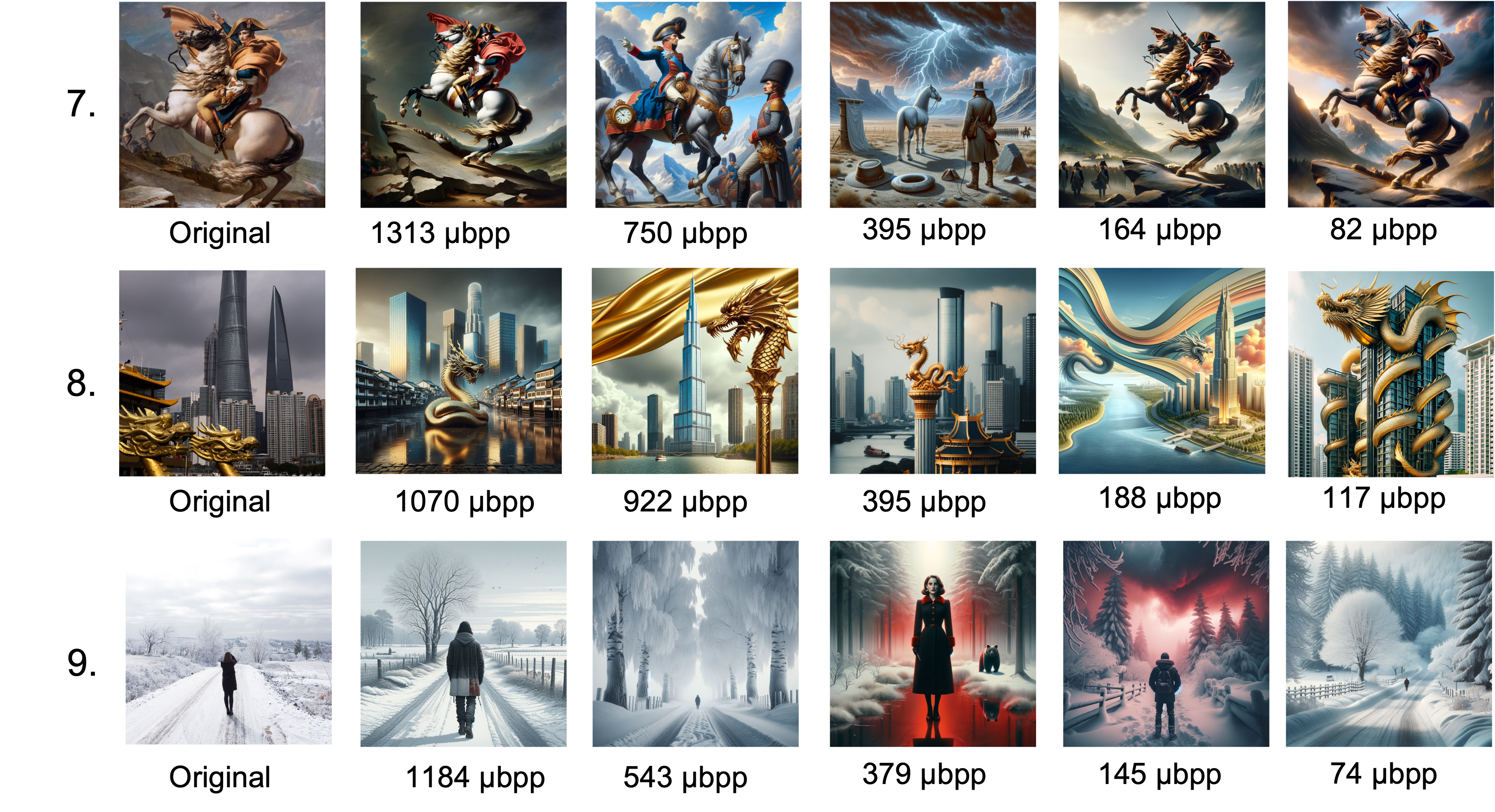}
  \vspace{-10pt}
  \caption{\textbf{More Compression Examples:} These examples show the usefulness of image-specific, variable-rate compression using fewer bits for more common images and gradual decline in quality in most examples at lower bitrates.}
  \label{fig:examples2}
\end{figure}

\begin{figure}[ht]
  \centering
  \includegraphics[width=\textwidth]{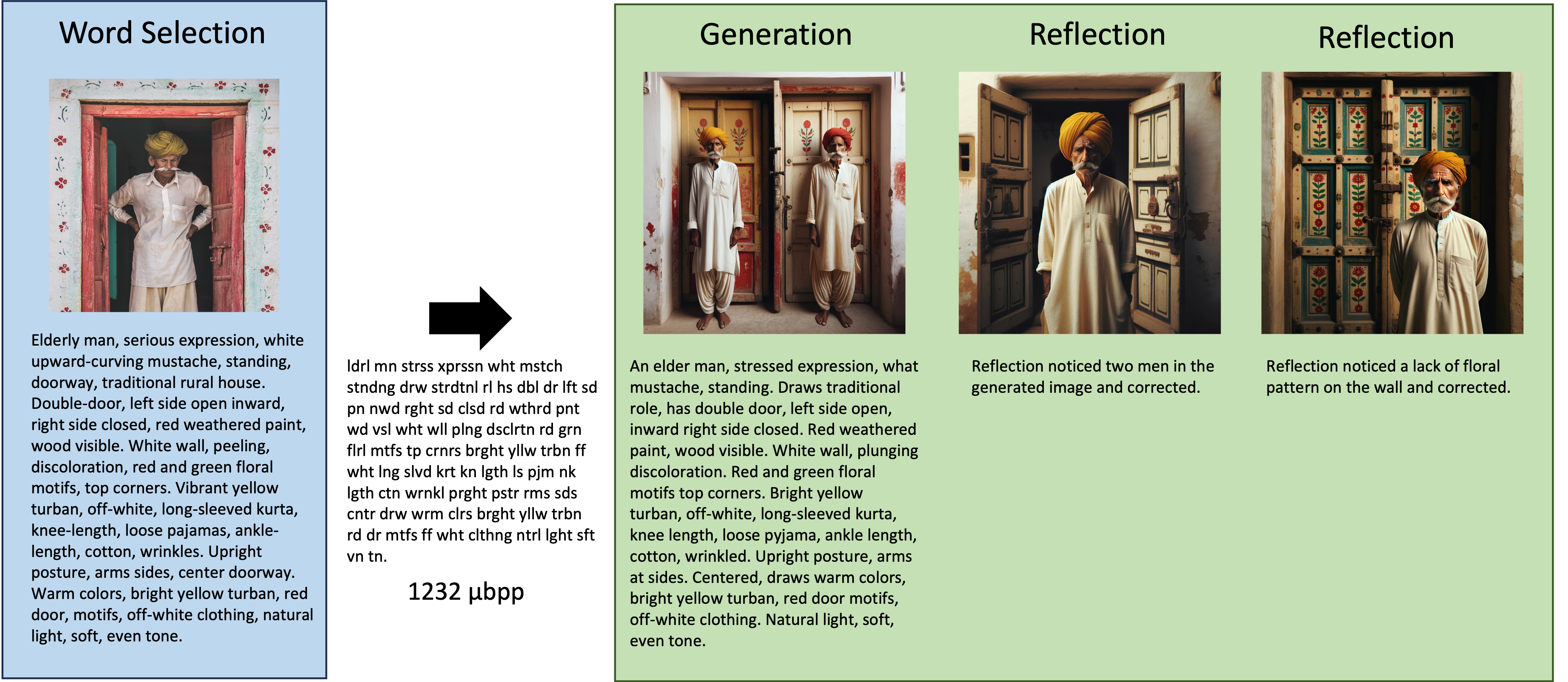}
  \vspace{-10pt}
  \caption{\textbf{Bearded Man:} An example of higher bitrates to demonstrate the effectiveness of reflection with sufficient context. Originally, the model produces two men and corrects for its mistake. Then, it has a regression on the floral pattern but identifies it and adjusts appropriately. It follows much of the detail in the uncompressed description.}
  \label{fig:beard}
\end{figure}

\begin{figure}[ht]
  \centering
  \includegraphics[width=\textwidth]{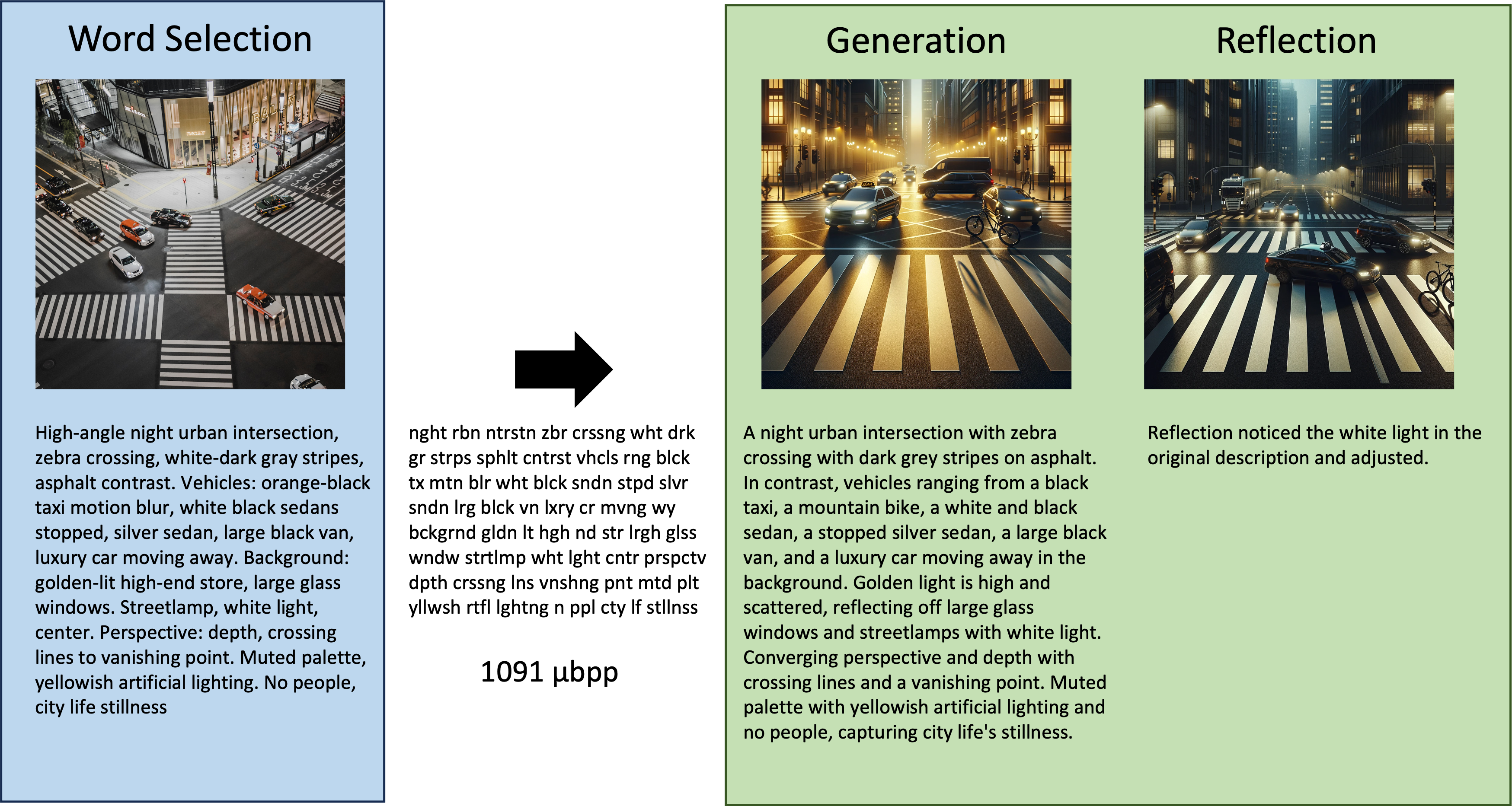}
  \vspace{-10pt}
  \caption{\textbf{City Block:} The model was able to recognize the issue with color of the street lighting, but it could not recognize the major difference in orientation. This is a common problem and also occurs with the Brooklyn Bridge example in Figure~\ref{fig:method}.}
  \label{fig:city}
\end{figure}

\begin{figure}[ht]
  \centering
  \includegraphics[width=\textwidth]{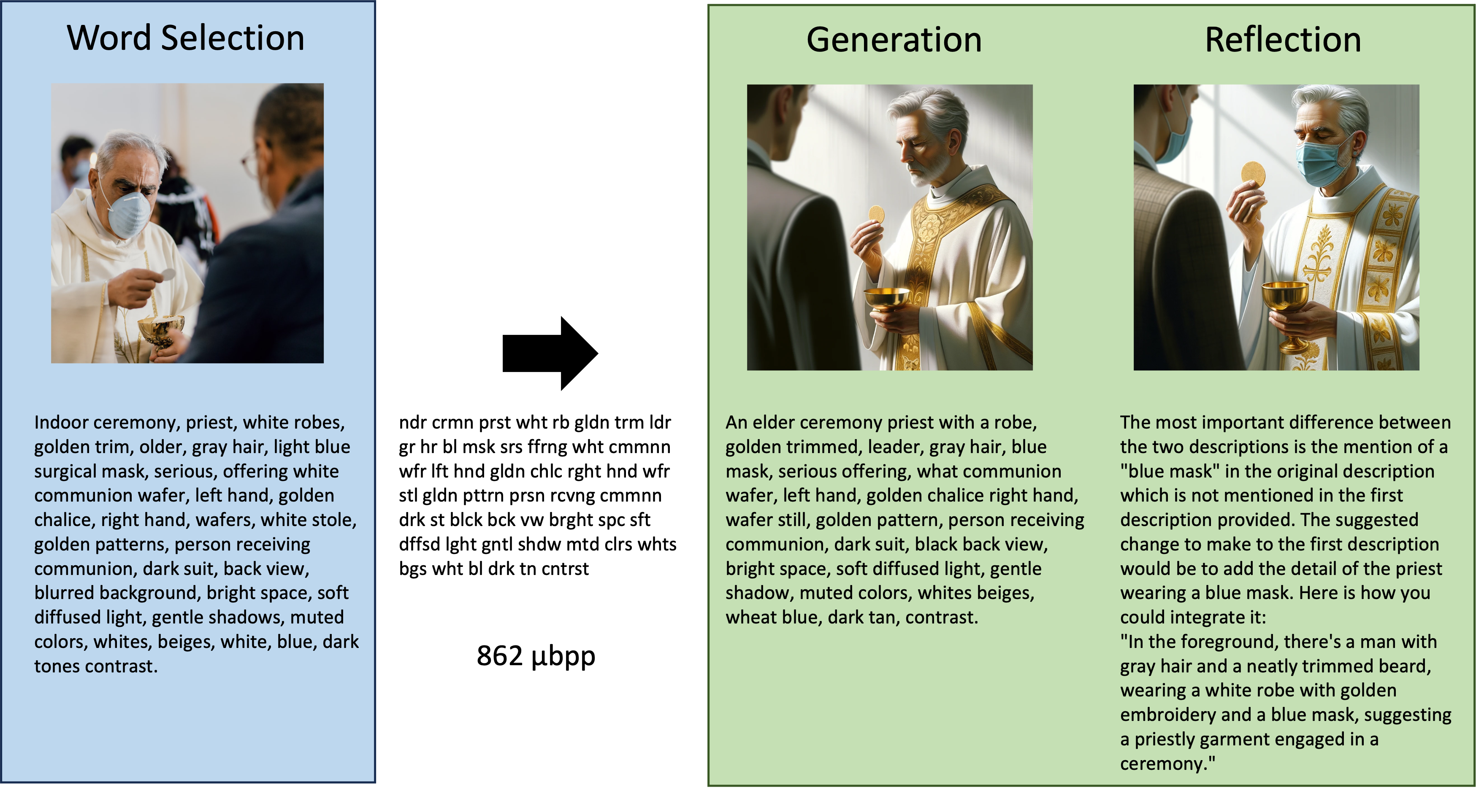}
  \vspace{-10pt}
  \caption{\textbf{Priest:} This example demonstrates the ability of DALL-E3 to make in place edits during reflection, although it still errors by adding masks to both the priest and the person receiving communion. It also incorrectly guesses the positions of the priest and other person, since there was no indication in the original description. }
  \label{fig:priest}
\end{figure}

\begin{figure}[ht]
  \centering
  \includegraphics[width=\textwidth]{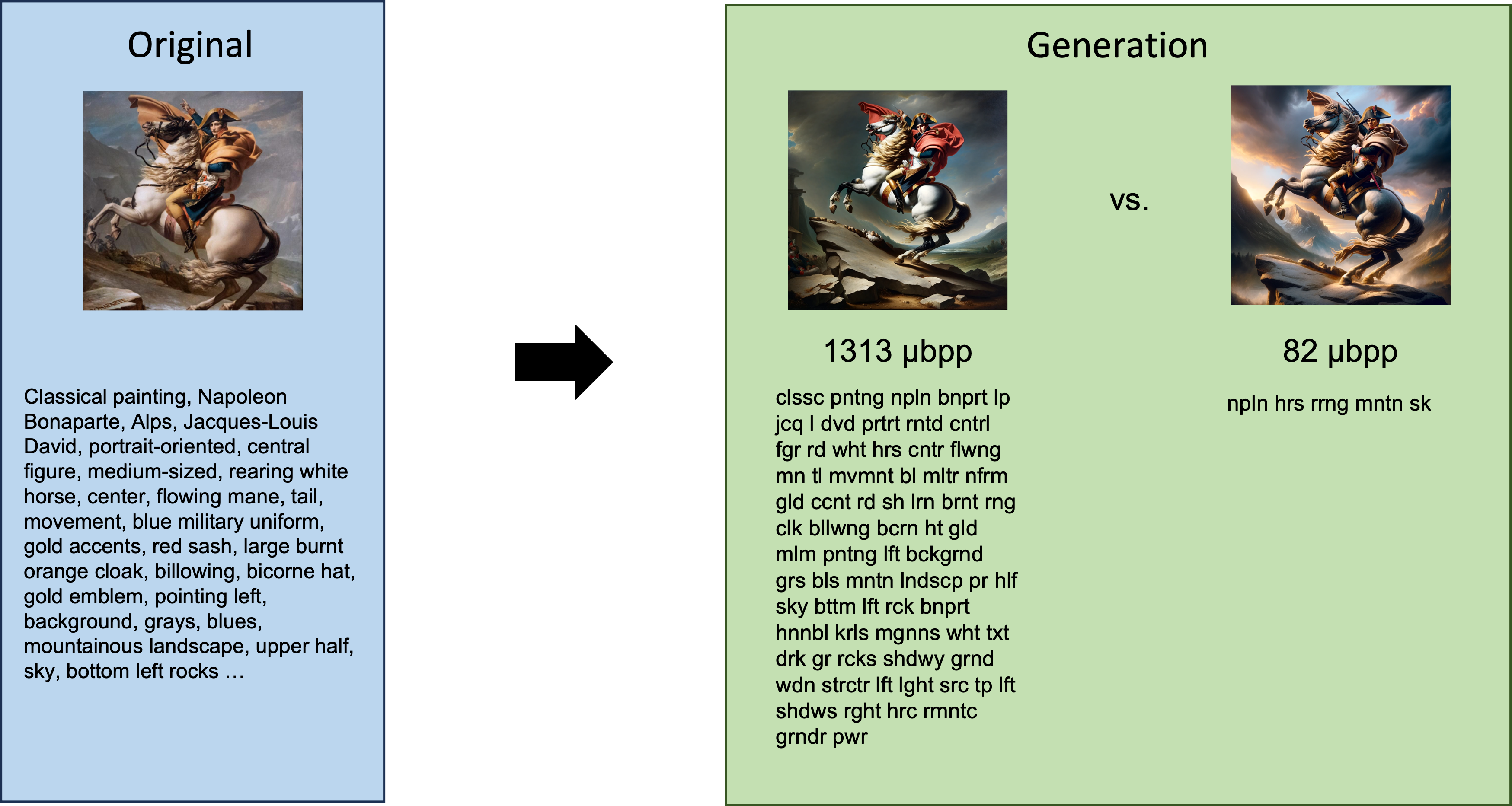}
  \includegraphics[width=\textwidth]{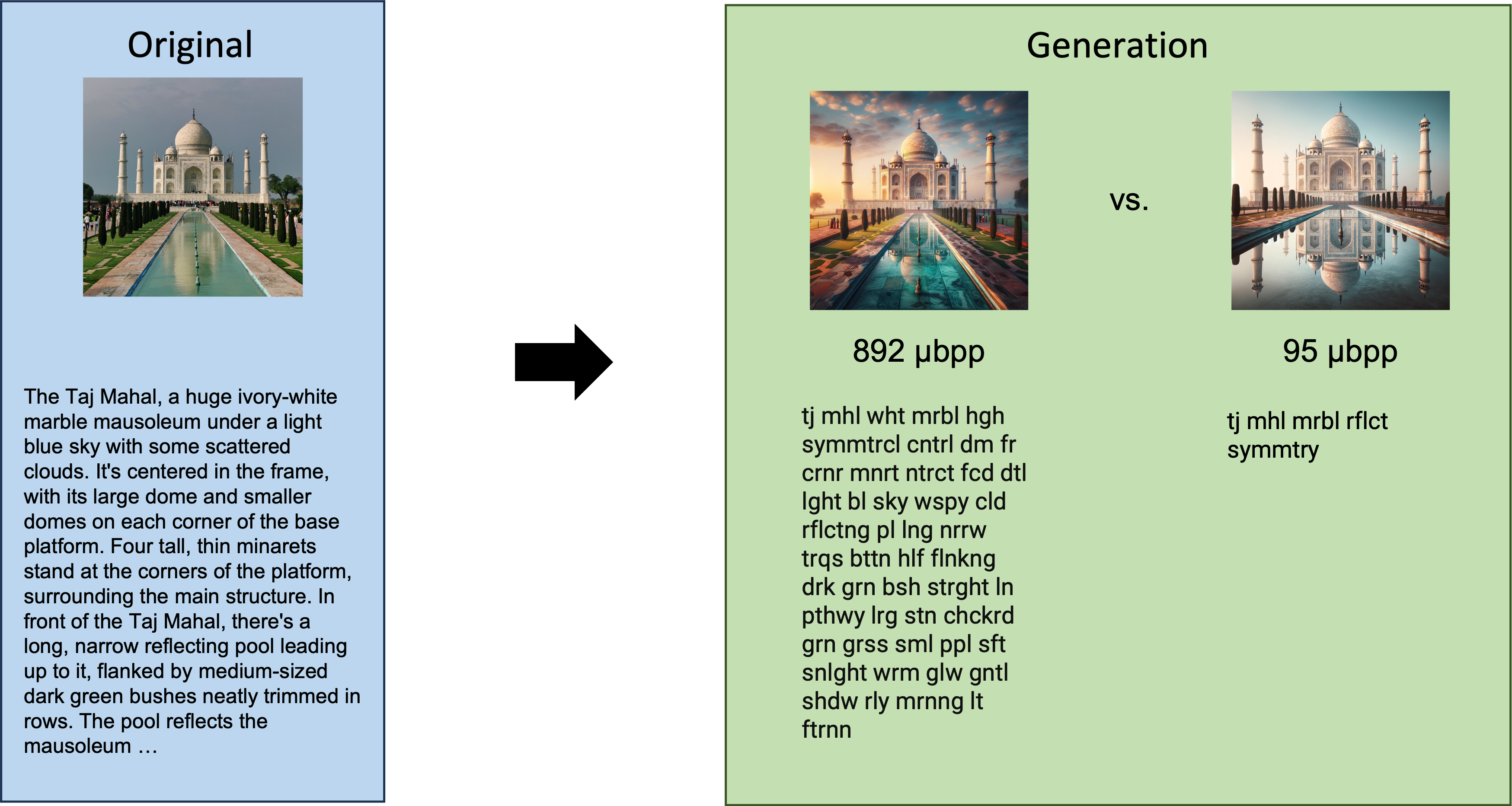}
  \vspace{-10pt}
  \caption{\textbf{Common Subjects:} For images of well-known subjects at standard angles, e.g., the Taj Mahal or Napoleon painting, very few words can produce accurate results. There is little appreciable increase in generated image accuracy from longer, more detailed descriptions. This phenomena could potentially be used to improve efficiency through a variable-rate compression algorithm.}
  \label{fig:taj}
\end{figure}

\clearpage

\section{Experiment Details}
\label{sec:experiment}

All the experiments are run with the GPT4 web interface, which can automatically make calls to the DALL-E3 API.
For simplicity, all images are square with a size of $1024 \times 1024$, which is a standard output size from DALL-E3.
The input images are cropped to this size before passing to GPT4-V.
If images are other sizes, then further super-resolution models can be used that can upsample the pixels using neural networks, or likely future decoder models will have better resolution support.
Most images are taken from the CLIC (Challenge on Learned Image Compression) dataset, since these images have already been filtered and selected for diversity.

This process is done manually, since at the time of publication DALL-E3 API is still under development and does not support consecutive API calls modifying previously generated images.
Without this support, the API cannot support reflection since each call is independent.
The encoder and the decoder are opened in different sessions to avoid any shared context.
Then, in the first session, the prompts below are used consecutively to describe the image, select the most important words, and then compress the characters in these words.
This compressed text is then passed to the second session, which decompresses the text and generates an image based on the description.
At the higher bitrates, there is enough context to perform reflection.

During reflection, the same description prompt that is used during encoding is used on the generated image to describe it.
Then, the uncompressed description available to the decoder is compared against this new description to select the most salient difference between the images.
This difference is passed again to DALL-E3 in the same session, and it can make adjustments to the previous image while attempting to minimize changes elsewhere in the image.
In this work, for examples that use reflection, the process continues for a fixed two iterations.
This is a hyper-parameter that balances quality and performance, and most images only have a few major potential issues after the initial generation.
Yet, in general, it is challenging to create an automated stopping condition for reflection.
This is in contrast with using reflection during code generation, where the stopping condition is determined by passing the test cases.

\subsection{Prompts}
\label{sec:prompts}

Below are the sets of prompts used for all examples in this paper for the encoding, decoding, and reflection tasks.
These were unchanged throughout the evaluation, and the current prompts were generated through manual trial-and-error.
The strength of the compression was determined by choosing the word count $K$ in the encoder Word Select prompt.
Given the inexactness of large language models, this word count is not always honored, yet GPT4-V typically has an error of less than $\pm 10\%$, and this behavior is actually desirable in many cases, since the model often only exceeds the limits with important words.

\subsubsection{Encode}
\textbf{Describe}: Can you describe this photo in as much detail as possible so that someone can recreate it based only on your description? Describe each object and its size in the image with small, medium, large, and huge. Describe the relative locations of all objects from the perspective of the viewer. Describe the colors in each object.

\textbf{Word Select}: This description will be used to regenerate an image. Can you compress this image description to $K$ words with the goal of selecting the most important words that humans would find relevant during the image reconstruction? These should be the most important words. Do not include helper words like prepositions or other unimportant words.

\textbf{Word Compress}: This description will be used to regenerate an image. Please remove all vowels and restrict to the following characters only: n, t, s, r, h, l, d, c, m, f, g, p, b, k, v. No punctuation is allowed. Remove plurals and uppercase letters.

\subsubsection{Decode}
\textbf{Word Decompress}: This is a description of an image that has been extremely compressed by removing vowels and punctuation. Keep in mind only these characters were allowed: n, t, s, r, h, l, d, c, m, f, g, p, b, k, v. Please decompress it to its original text.

\textbf{Generate}: Please generate a square image based on this description by following all of the details.

\subsubsection{Reflect}
\textbf{Describe}: Can you describe this photo in as much detail as possible so that someone can recreate it based only on your description? Describe each object and its size in the image with small, medium, large, and huge. Describe the relative locations of all objects from the perspective of the viewer. Describe the colors in each object.

\textbf{Compare}: Please compare the image description above to the original description below and highlight the most important difference between the two. Format this difference into a suggested change to make to the description above to make it more like the original description below.

\textbf{Generate}: Please keep the exact same image but make the following change:

\section{Reflection}
\label{sec:reflection}

Reflection is the process of iterative development, which mirrors the human generative process in writing, painting, and other creative tasks.
Variants of reflection within language models have been explored in other fields to improve the quality of generative models.
For example, Reflexion~\cite{shinn2023reflexion} uses it to significantly improve the code generation and achieve state-of-the-art results on the HumanEval task.
It first produces candidate code and executes it, analyzes its output or error messages, and then iterates until the code passes a set of test cases or reaches a maximum number of iterations.
Our work applies a similar method that generates, analyzes, and iteratively improves images.
Specifically, it compares the decoded text description of the original image with a new description of the generated image and highlights the major semantic differences.
Then, it suggests the most important change following the prompts in the appendix, and regenerates the image with this change.

Our use of reflection was further motivated by the high descriptive strength of the GPT4-V compared to the generative strength of DALL-E3.
This is similar to the classic N=NP problem, which likely suggests it is easier to evaluate that a solution is correct than generate a solution itself. 
Therefore, the reflection process can make up for relatively poor performance of the decoder through iterative analysis and generation on the initial design.
In practice, this process is currently limited by the ability of the decoder to isolate these changes and can lead to regressions during reflection.
Sometimes iterations can significantly change the previous image or undo changes made during previous reflection iterations.
Overall, however, reflection typically leads to the most serious issues with the generated image being fixed typically within a few iterations.

\end{document}